\documentclass{article}
\usepackage[final,sglblindworkshop]{neurips_2026}
\workshoptitle{AI4Mat Workshop at NeurIPS 2026}

\usepackage[utf8]{inputenc}
\usepackage[T1]{fontenc}
\usepackage{hyperref}
\usepackage{url}
\usepackage{booktabs}
\usepackage{amsmath}
\usepackage{amssymb}
\usepackage{microtype}
\usepackage{xcolor}
\usepackage{graphicx}
\usepackage{caption}
\usepackage{array}

\title{DFT-Agent Bench: Scaffold Effects and\\
Execution Provenance in Autonomous DFT}

\author{
Shijie Liu$^{1}$ \quad
Jin Yang$^{2,3,4}$ \quad
Weilin Wan$^{5}$ \\
Jingtao Han$^{6}$ \quad
Jia LI$^{6}$\thanks{Corresponding author:
\texttt{jia@projectnumina.ai}.} \quad
Tianqi Deng$^{2,3}$\thanks{Corresponding author:
\texttt{dengtq@zju.edu.cn}.} \\[0.5em]
$^{1}$Shanghai Jiao Tong University \\[0.3em]
$^{2}$State Key Laboratory of Silicon and Advanced Semiconductor Materials \\
and School of Materials Science and Engineering, Zhejiang University, \\
Hangzhou, Zhejiang 310027, China \\[0.3em]
$^{3}$Institute of Advanced Semiconductors, \\
ZJU-Hangzhou Global Scientific and Technological Innovation Center, \\
Zhejiang University, Hangzhou, Zhejiang 311200, China \\[0.3em]
$^{4}$School of Physics, Zhejiang University, Hangzhou 310027, China \\[0.3em]
$^{5}$Fudan University \\
$^{6}$Project Numina \& WarpDrive Intelligence \\[0.5em]
\texttt{sjtu7181681@sjtu.edu.cn} \quad
\texttt{jiny25@zju.edu.cn} \\
\texttt{wlwan23@m.fudan.edu.cn} \quad
\texttt{jingtao@projectnumina.ai} \\
\texttt{jia@projectnumina.ai} \quad
\texttt{dengtq@zju.edu.cn}
}
\begin{document}
\maketitle

\begin{abstract}
Large language models are increasingly deployed as autonomous agents for
first-principles materials simulation. We report empirical findings from
\textbf{DFT-Agent Bench}, a benchmark of 200+ Quantum ESPRESSO tasks spanning
relaxation, electronic structure, equation-of-state, and elastic-tensor
pipelines over multiple material tiers. Each task carries a QE-derived ground
truth with public-reference cross-checks where available, and a
\emph{provenance audit} checks whether the required QE
execution contract was met. Under one execution harness, we compare a
near-bare ReAct baseline with a bundled v5.1 scaffold on two models. The
complete scaffold improves bands+DOS by $+0.08$/$+0.11$ and $d$-electron
vc-relax by $+0.17$ on both models, while high-scoring relaxation and
equation-of-state cells leave little room for improvement. The elastic result
is model-dependent ($0.18\to0.40$ for \texttt{deepseek-v4-pro} and
$0.62\to0.59$ for \texttt{claude-opus-4-8}). A fixed 100-task analysis with
retrieval disabled supports the aggregate pattern: the fully frozen DeepSeek
comparison gains $+0.17$, while the Opus comparison gains $+0.10$ against a
historical pool baseline ($p{<}0.001$). The provenance gate catches incomplete
or unmarked executions that can receive high rubric scores; it is an
execution-contract check, not a test for intentional fraud. These results
characterize the evaluated DFT workflows. The anonymous review artifact contains
the frozen diagnostic slice, prompts, scorer, and solver revisions:
\url{https://anonymous.4open.science/r/231e31re91287h/README.md}.
\end{abstract}

\section{Introduction}
Agentic LLMs now drive density-functional-theory (DFT) workflows end to end,
preparing inputs, launching planewave codes, and parsing
results~\citep{liu2025vaspilot,liu2025masgent,xia2025vaspagent,kumar2026elagente,pham2026chemgraph}.
Each system wraps its model in more scaffolding (structured tool APIs,
retrieved domain knowledge, verifier loops, multi-agent decomposition).
Each is typically reported in isolation and scored by a rubric on its final
report. Two questions are rarely answered together: \emph{how much does this
engineering actually help} relative to simply prompting a frontier model, and
\emph{can we trust the reported numbers} as the product of a real
calculation? General agent
benchmarks~\citep{jimenez2024swebench,liu2024agentbench,mialon2024gaia,chan2024mlebench,chen2025scienceagent}
test task outcomes; in physics, provenance is part of correctness. Recent
scaffold-versus-model analyses also show that outcome and process can diverge
and that model choice and scaffold choice need to be measured separately
\citep{riosgarcia2026aiscientists}. We therefore treat the present study as a
domain-specific controlled comparison, rather than a claim about agent
engineering in general.

We make three contributions. (i) \textbf{DFT-Agent Bench}: 200+ Quantum
ESPRESSO (QE)~\citep{giannozzi2009qe,giannozzi2017qe} tasks across four task
types and multiple material tiers, with QE-derived ground truth. (ii) A
\textbf{provenance-audited scorer}: a four-flag execution
audit that sets the score to zero when the required execution contract fails. (iii) A \textbf{controlled comparison of the baseline and engineered
flagship} using two models under one harness. Across this evaluated pool, the
complete scaffold improves selected low-baseline cells, while high-scoring
cells show smaller gains. A frozen 100-task analysis with retrieval disabled
provides a cleaner diagnostic of the aggregate comparison. Component-level
attribution and general scientific reasoning remain outside this package-level
evaluation.

\section{DFT-Agent Bench and audited scoring}
\paragraph{Tasks and ground truth.}
\looseness=-1
Each task is a pair (material tier $\times$ task type), with the two axes
decoupled: the tier sets chemical difficulty, the task type sets
pipeline length. Tiers ascend from M1 (elemental and binary solids with simple crystal
structures, including face-centered-cubic/body-centered-cubic/%
hexagonal-close-packed elemental metals, diamond, and zincblende)
through M2 (large unit-cell oxides/complex compounds such as perovskites and
skutterudites) to M3 (magnetic transition-metal $d$-electron systems with
both ferromagnetism and antiferromagnetism, needing spin-polarized
calculations). The released set consists of the two
stable tiers M1+M2, 245 unique task IDs over four pipelines of increasing length:
\textbf{T1} a single variable-cell relaxation; \textbf{T2}, a
self-consistent-field (SCF) calculation followed by non-self-consistent-field
(NSCF) calculations of the band structure and density of states (DOS);
\textbf{T3}, an equation-of-state (EOS) calculation that fits energy--volume
data to the Birch--Murnaghan equation of state to extract the equilibrium
volume and bulk modulus; and \textbf{T4}, calculation of the elastic tensor
$C_{ij}$ by applying finite strains to a relaxed reference structure. We
additionally conduct an exploratory evaluation of \textbf{T1} and \textbf{T2}
at the more challenging M3 tier. The ground truth is produced by our own
converged QE runs; where public reference values are available, the package
reports cross-validation at approximately $1\%$ relative tolerance against
established
references~\citep{bosoni2024acwf,prandini2018sssp,jain2013materialsproject,dunn2020matbench,choudhary2024jarvis}.
All QE calculations used the PBE exchange--correlation
functional~\citep{perdew1996pbe} and a fixed set of GBRV ultrasoft
pseudopotentials~\citep{garrity2014gbrv}. For M3 materials, the benchmark
specifies the magnetic ordering and initial magnetic moments. Each task
is graded by a task-type-specific, graded-relative rubric (full credit within
a fraction of the tolerance, decaying to zero at it): a relative error on
scalars such as the bulk modulus, an eigenvalue-based relative error on
$C_{ij}$, and a curve RMSE on the energy--volume and stress--strain profiles.

\paragraph{Audited scoring.}
\looseness=-1
Beyond the rubric, every run passes through a four-flag execution audit:
\texttt{qe\_invoked} (a QE binary actually ran), \texttt{job\_done} (the full
methodology grid completed, every point carrying QE's completion marker),
\texttt{qe\_wallclock\_ok} (a physically plausible wall-clock time), and
\texttt{no\_network\_access} (no egress).
If a run fails any of these checks, we assign it a score of zero. The audit
detects failures that numerical scoring alone can miss. For example, a
seven-point EOS curve may satisfy the numerical tolerance even though QE
successfully completed only three of the seven required calculations. The
audit therefore verifies execution conditions; this manuscript does not claim
that the four flags alone form a complete numerical result-to-output chain.
Only a small fraction of runs fail the audit (Section~\ref{sec:results}). All
scores reported below incorporate this rule.

\paragraph{Run-pool hygiene.}
\looseness=-1
Because the audited pool accumulated while the harness was hardened, the final
analysis snapshot applies comparability filters uniformly to both agents.
\emph{Infrastructure runs} are runs with no assistant turn or a provider-side
API outage. \emph{Legacy runs} precede the current evaluator/corrected elastic
rubric or the current T3/T4 prompt epoch (the snapshot records epoch
$\geq$ 2026-06-17). The source snapshot records 183 rows re-scored under the
current evaluator and a read-only T2 check with 370/370 extracted rows and no
material drift. These metadata and epoch rules remove $1{,}140$ infrastructure
runs and $125$ legacy-prompt/evaluator runs. Agent-level failures after
evaluation begins---including incomplete methodology grids, rejected
submissions, provenance failures, and budget exhaustion---remain in the
denominator and receive zero under the audited scorer.

The v5.1 package includes structured QE tools, input linting, a domain
knowledge layer, an SCF convergence sub-agent, and a verifier gate. It also
ships an experience store. The main-pool analysis records retrieval as
inactive, while the frozen analysis explicitly disables it with
\texttt{--no-experience}; the matched A/B is therefore the only direct test of
this component. When enabled, the artifact's store persists methodology
records under \texttt{data/experience.jsonl} with a parallel embedding index;
the frozen artifact omits these files because retrieval is off.

\paragraph{Agents.}
\looseness=-1
We compare two reference solvers under one harness contract. \textbf{v1} is a
near-bare ReAct loop with four generic tools (write file, run shell, read output,
submit result), a faithful ``just prompt a frontier model'' baseline.
\textbf{v5.1} is an engineered flagship: structured QE tools with input linting, a
domain-knowledge layer, an SCF convergence sub-agent, and a verifier gate. The
comparison is between these complete packages; it is not a component-level
ablation. Both run on two frontier models (\texttt{deepseek-v4-pro}, \texttt{claude-opus-4-8})
under identical limits: up to 100 tool-calling turns and a $7200$\,s wall-clock
budget per task, against a GPU build of Quantum ESPRESSO~7.5 in an isolated
container (caps raised for the frozen batches of Finding~4).

\section{Results}\label{sec:results}
\begin{table}[t]
\centering
\caption{Historical/descriptive pooled summary of per-cell raw and
provenance-gated scores, v1 vs.\ v5.1, \emph{pooled over both models}: best
score per retained historical model--task record. The $n$ values are retained historical pool entries, not unique released
task IDs: the six M1+M2 rows sum to 552 entries per arm because repeats and
task versions are retained, whereas the released set has 245 unique IDs. Raw and gated nearly coincide except M1 vc-relax, where
flagship contract misses are voided ($0.97$ vs $0.91$; Finding~5). The pooled
elastic row averages a bimodal split ($0.18\!\to\!0.40$ deepseek vs
$0.62\!\to\!0.59$ opus; Fig.~\ref{fig:compare}); M3 rows are per-model in
Fig.~\ref{fig:tierheat}. M2 multi-step cells ($n{=}10$) omitted.}
\label{tab:cells}
\footnotesize
\begin{tabular}{llcccc}
\toprule
& & \multicolumn{2}{c}{raw} & \multicolumn{2}{c}{gated} \\
\cmidrule(lr){3-4}\cmidrule(lr){5-6}
Tier \& task & $n$ (v1/v5.1) & v1 & v5.1 & v1 & v5.1 \\
\midrule
M1 \,vc-relax    & 116/116 & 0.96 & 0.97 & 0.96 & 0.91 \\
M1 \,bands+DOS   & 116/116 & 0.78 & 0.91 & 0.78 & 0.91 \\
M1 \,EOS         & 104/104 & 0.97 & 0.98 & 0.95 & 0.97 \\
M1 \,elastic     & 104/104 & 0.45 & 0.51 & 0.43 & 0.51 \\
M2 \,vc-relax    & \phantom{0}56/\phantom{0}56 & 0.94 & 0.95 & 0.93 & 0.94 \\
M2 \,bands+DOS   & \phantom{0}56/\phantom{0}56 & 0.77 & 0.78 & 0.77 & 0.77 \\
M3 \,vc-relax    & \phantom{0}30/\phantom{0}30 & 0.78 & 0.93 & 0.76 & 0.93 \\
M3 \,bands+DOS   & \phantom{0}30/\phantom{0}30 & 0.68 & 0.70 & 0.68 & 0.70 \\
\bottomrule
\end{tabular}
\end{table}

\begin{figure}[t]
\centering
\includegraphics[width=0.67\linewidth]{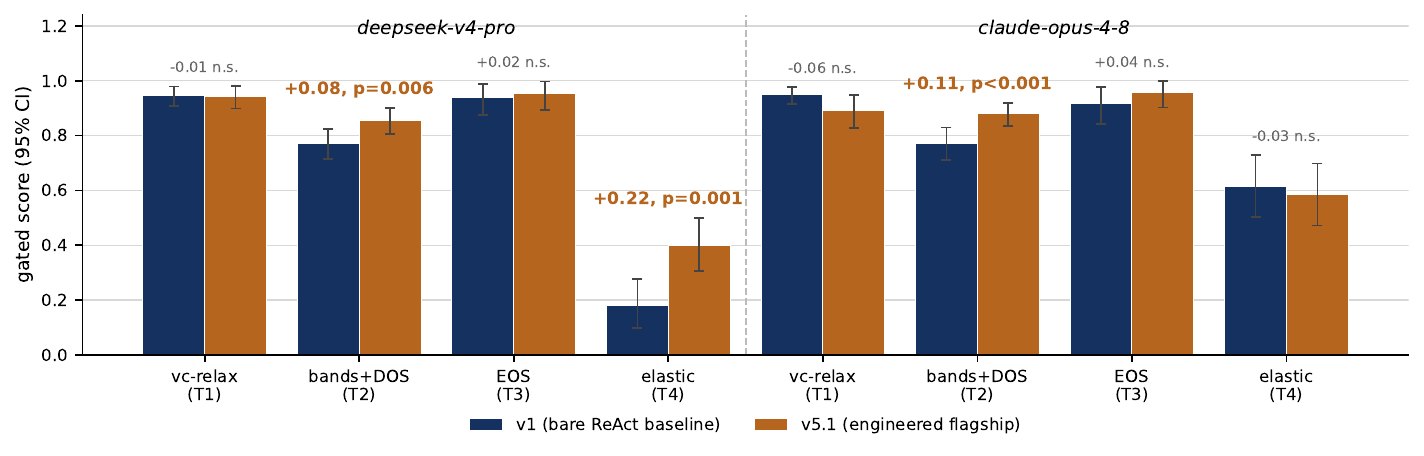}
\caption{Model-matched v1 vs.\ v5.1 with 95\% task-clustered bootstrap CIs
(M1+M2 common tasks; no cross-model pooling). Three cells separate: bands+DOS
on both models and elastic on \texttt{deepseek-v4-pro} only; vc-relax and EOS
are flat, and the opus \emph{bare baseline} already reaches $0.62$ on
elastic (Finding~3).}
\label{fig:compare}
\end{figure}

\begin{figure}[t]
\centering
\includegraphics[width=0.66\linewidth]{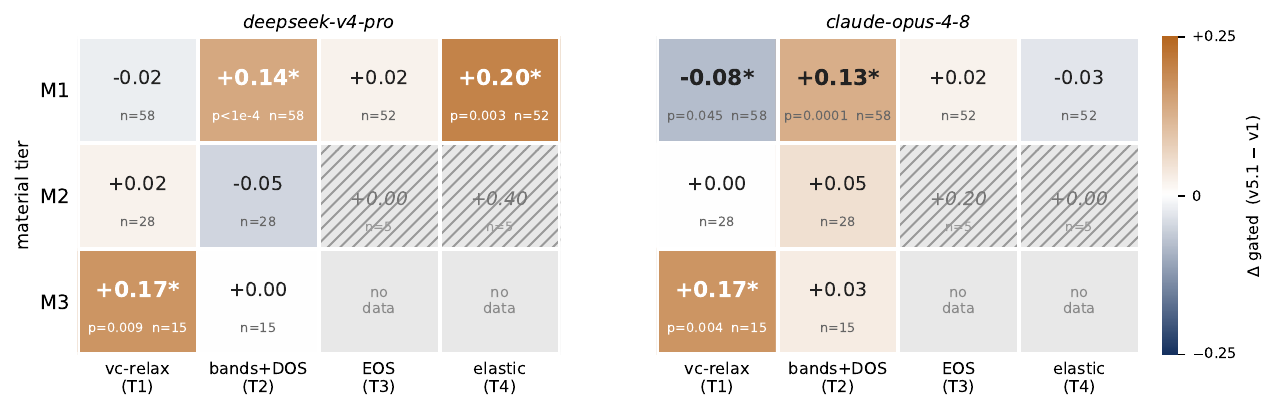}
\caption{\textbf{Observed package gains in selected lower-scoring cells, per
model}: $\Delta=\,$gated(v5.1)$\,-\,$gated(v1) within each model over paired
tasks. M1 bands+DOS gains are $+0.14$/$+0.13$ and M3 vc-relax gains are
$+0.17$/$+0.17$; elastic separates on deepseek only ($+0.20$; Finding~3).
The one negative cell, opus M1 vc-relax ($-0.08$), is dominated by audit
contract misses (Finding~5). Hatched: $n{<}10$, too thin to trust.}
\label{fig:tierheat}
\end{figure}

\paragraph{Finding 1: the complete v5.1 package helps selected DFT regimes.}
\looseness=-1
Head-to-head (Fig.~\ref{fig:compare}), the agents are within noise wherever
the baseline is already strong: vc-relax and EOS on both models and every M2
cell. The one exception, opus vc-relax ($-0.06$), traces to audit contract
misses (Finding~5). In the evaluated configurations where the baseline has
lower scores, the complete package improves bands+DOS
$+0.08$ ($p{=}0.006$) on \texttt{deepseek-v4-pro} and $+0.11$ ($p{<}0.001$)
on \texttt{claude-opus-4-8}, largely the metal band-path protocol the bare
baseline often misses. On the $d$-electron tier M3, vc-relax gains $+0.17$ on both
models ($p{=}0.009$/$0.004$, Fig.~\ref{fig:tierheat}). These observations
support a scoped package-level comparison; they do not identify which bundled
component produced the changes.

\paragraph{Finding 2: bounded scores and stochasticity limit interpretation.}
On the easier relaxation and EOS cells, both agents score around ${\sim}0.9$
or above, so the bounded rubric leaves little numerical headroom. This is
score saturation within this benchmark, not evidence that the models have
reached a general scientific capability ceiling. Single runs are noisy: over
the 172 bands+DOS configurations with repeated trials, the
within-configuration standard deviation averages $0.15$, larger than the agent
gaps. Per-task differences therefore need $k\!\ge\!3$ repeats; the cell-level
gaps stay readable by aggregating $30$--$116$ pairs each.

\paragraph{Benchmark familiarity.}
Some structures, reference values, and workflow conventions derive from
long-public resources, including the Materials Project and published DFT
literature~\citep{jain2013materialsproject}. We cannot rule out that task
content or reference data appeared in model pretraining. High scores therefore
measure performance on these workflows under the shared harness; they should
not be read as clean evidence of out-of-distribution scientific discovery or
independent rediscovery of the reference knowledge.

\paragraph{Finding 3: the elastic comparison is model-dependent.}
\looseness=-1
The elastic tensor is the hardest task and the one cell where the two models
disagree about engineering. \texttt{deepseek-v4-pro}'s bare baseline scores
$0.18$ after a systematic stress-sign convention error in its self-written
fitting code (correct magnitudes, inverted signs).
The flagship reaches $0.40$ under its deterministic strain protocol
($p{<}0.001$). \texttt{claude-opus-4-8}'s bare baseline makes no such error and reaches
$0.62$; the flagship adds nothing ($-0.03$, n.s.), so the pooled elastic row
of Table~\ref{tab:cells} averages two different stories. The
flagship's remaining elastic failures are budget exhaustion without a
complete strain grid.
The deepseek result is consistent with the v5.1 protocol checks avoiding this
stress-sign failure, but the bundled comparison cannot establish that component
as the cause. Figure~\ref{fig:compare} contains four model-matched M1+M2
cells for each model, hence eight paired tests; a Bonferroni correction at
$\alpha=0.05$ has cutoff $0.00625$. The displayed deepseek bands+DOS value
($p{=}0.006$) is rounded to three decimals, and the unrounded value and test
script are not in the package, so we leave its corrected classification open.
The displayed $p{<}0.001$ and $p{=}0.001$ results are below that cutoff; the M3
values ($p{=}0.009$/$0.004$) are exploratory and outside this eight-test family.

\paragraph{Finding 4: a fixed, retrieval-off diagnostic supports the
package-level comparison.}
\looseness=-1
We froze a 100-task slice and re-ran both agents on fixed code with retrieval
disabled (\texttt{--no-experience}): the flagship on both models, the baseline
on deepseek. The anonymous artifact documents the fully paired DeepSeek arm and
states that the slice was defined in advance for diagnosis. The flagship lands
at model parity ($0.88$/$0.88$; per-task gated means, failures as zero). With
both sides frozen (deepseek), the paired gain is $+0.17$
($0.71\!\to\!0.88$, $p{<}10^{-4}$, $100$/$100$ pairs). The pool's best-of-many
attempts flatter the baseline by $+0.08$ yet leave $+0.09$ ($p{<}0.001$),
matching opus versus its pool baseline ($+0.10$, $p{<}0.001$). An A/B against
matched same-code retrieval-\emph{on} runs finds no detectable contribution
($\Delta{=}{+}0.03$ favoring \emph{off}, n.s., $n{=}21$). The artifact does not ship the diagnostic label file or its selection
predicate; we therefore use the aggregate comparison only and omit subgroup
interpretation.
A controlled DFT+$U$ diagnostic dissects the two remaining near-zeros, both
antiferromagnetic oxides. CoO's miss is bench-side: the ground truth is
itself the spurious $U$-free PBE cell distortion that $+U$ removes, so the
agent's physically sound refusal to submit it scores zero.
Cr$_2$O$_3$'s miss is agent-side: a converged $U$-free run does reach the
AFM insulating state ($1.2$\,eV gap); the submitted run missed it, scoring
its DOS metallic ($0.10$).

\paragraph{Finding 5: provenance auditing catches execution-contract failures.}
\looseness=-1
The audit voids $1.6\%$ of scored runs for both agents (11/706 baseline,
15/942 flagship): overwhelmingly correct-\emph{looking} answers (20 of 26
at raw $\ge 0.85$), voided for an incomplete grid or a missing completion
marker. The gate cuts both ways: six flagship opus vc-relax runs with perfect
rubric scores were voided for declaring the wrong output log (the raw-vs-gated
gap on M1 vc-relax, Table~\ref{tab:cells}); the frozen batches added five
(one flagship, four baseline). Thus the gate enforces execution contracts and surfaces runs that fail those
checks; it is not an intent detector and does not by itself trace every
submitted scalar to an audited output. Without it, roughly $2\%$ of scored
runs in this pool would bypass the execution gate.

\section{Related Work}
\looseness=-1
Recent systems wrap a frontier model in a domain harness for
planewave or materials workflows (DREAMS~\citep{wang2025dreams},
VASPilot~\citep{liu2025vaspilot},
Masgent~\citep{liu2025masgent}, VaspAgent~\citep{xia2025vaspagent},
ElAgent~\citep{kumar2026elagente}, ChemGraph~\citep{pham2026chemgraph}, and
others~\citep{soleymanibrojeni2025genius,zhang2026matclaw,tritondft2026}).
Each is engineered in isolation and rarely quantifies what its scaffolding
adds. Recent cross-domain work evaluates model and scaffold contributions
jointly and reports that correct outputs can coexist with scientifically weak
processes~\citep{riosgarcia2026aiscientists}. Our narrower contribution is a
DFT-specific comparison of two complete solver packages under one harness,
with execution checks and failure diagnoses.
General~\citep{jimenez2024swebench,liu2024agentbench,mialon2024gaia,chan2024mlebench}
and scientific~\citep{chen2025scienceagent,siu2025mattools,wang2024scicode}
agent benchmarks score outcomes; for first-principles work a
correct-looking report is no evidence of a valid isolated run. Our audit checks
the execution contract and makes that distinction explicit.

\section{Discussion}
\looseness=-1
Within this benchmark, the hardest cells remain unsolved for every
agent--model pair: the elastic tensor tops out at $0.40$--$0.62$ and M3
bands+DOS near $0.7$. The complete v5.1 package recovers performance in some
lower-scoring configurations, but does not move this observed frontier. The
experience-store A/B does not show a detectable contribution in the tested
comparison. These are package-level observations under one harness, and the
bundled components are not individually attributed.

\emph{Limitations.} The analysis uses two custom solver packages and two
models. The historical pool is heterogeneous: prompts and evaluators changed,
repeat counts are unequal, and the table reports best-of-many values for some
model--task cells. Such values are not equivalent to a balanced fixed-repeat
estimate, and agent stochasticity (within-configuration standard deviation
about $0.15$ in the repeated bands+DOS pool) limits task-level conclusions.
The frozen comparison is cleaner but does not cover every cell equally; the
opus arm uses a pool-drawn baseline while the deepseek arm was re-run frozen.
M2 multi-step and M3 coverage is partial. We do not have a full component
ablation or a mature external scaffold baseline, and benchmark materials and
reference values are public enough that familiarity or training-data
contamination cannot be excluded. Future work should document subgroup rules
before analysis, balance repeats, compare additional scaffolds, and test
component-level causality.

\newpage
\appendix
\section{Reproducibility appendix}
\label{app:reproducibility}
The anonymous review artifact was checked at
\url{https://anonymous.4open.science/r/231e31re91287h/README.md}. It contains
\texttt{README.md}, task JSONs, prompt templates, the harness, solver sources,
the frozen score table, and the analysis script used for the matched diagnostic.

\paragraph{Tasks and frozen slice.}
The workshop source defines 245 unique stable M1+M2 task IDs crossed with
T1--T4; the paper also reports exploratory M3 T1/T2 tasks. T1 is variable-cell
relaxation, T2 is SCF+NSCF bands/DOS, T3 is an equation-of-state fit, and T4
is a finite-strain elastic tensor. The artifact is a separate 100-task
diagnostic slice: \texttt{results/frozen\_matched\_scores.csv} contains one
gated score per agent and task, and \texttt{analysis/frozen\_matched.py}
selects batches \texttt{v1-ds-frozen} and \texttt{jobspec-2026-07-05}, keeps
the maximum score per task, and checks for 100 paired tasks. Its README says
the slice was defined in advance and is not a random sample. The script
requires a 73-task headroom-list input, but that list and its predicate are not
included; subgroup-specific claims are therefore omitted from the main
findings. Ground truth uses converged QE calculations and, where public values
are available, cross-validation at approximately $1\%$ relative tolerance
against references including the Materials Project~\citep{jain2013materialsproject}.

\paragraph{Prompts, agents, and tools.}
The verified frozen files include
\url{solvers/v1/agent/prompts/system.txt} and the v5.1 prompt files
\url{solvers/v5_1/agent/prompts/executor.md},
\url{solvers/v5_1/agent/prompts/strategist.md},
\url{solvers/v5_1/agent/prompts/verifier.md}, and
\url{solvers/v5_1/agent/prompts/convergence_subagent.md}.
The v5.1 manifest is \texttt{solvers/v5\_1/manifest.yaml}, version 0.5.0;
structured QE tools are under \texttt{solvers/v5\_1/agent/tools/structured/},
input linting is \texttt{agent/tools/qe\_input\_lint.py}, verification is
\texttt{agent/tools/structured/verify\_run.py}, the convergence sub-agent is
\texttt{agent/subagents/convergence.py}, and domain skills are under
\texttt{agent/skills/}. The artifact records solver revisions
\texttt{82feeec} (v1), \texttt{10f9283} (v5.1), and benchmark revision
\texttt{6aa682bd0d5e61598118b71984ea2f85711483e7}. The artifact manifest is a
review export for the frozen T1/T2 slice; it is not a manifest for the full
historical T3/T4 pool.

\paragraph{Run selection and scoring.}
The workshop snapshot annotation records the source paths
\texttt{repro/build\_corrected\_snapshot.py},
\texttt{analysis/regen\_results\_corrected.py}, and
\texttt{repro/rescore\_t2\_readonly.py}. Its filters define infrastructure
runs as having no assistant turn or a provider-side API outage, and legacy runs
as predating the current evaluator/corrected elastic rubric or the current
T3/T4 prompt epoch ($\geq$ 2026-06-17). The snapshot records 183 rows
re-scored under the current evaluator and a 370/370 read-only T2 extraction
check. These rules remove 1,140 infrastructure and 125 legacy-prompt/evaluator
runs; failures after evaluation begins remain zero-score observations. In the
verified artifact, \texttt{harness/lib/audit.py} gates on
\texttt{qe\_invoked}, \texttt{job\_done}, \texttt{qe\_wallclock\_ok}, and
\texttt{no\_network\_access}; \texttt{paths\_in\_workdir} and
\texttt{files\_nonempty} are additional integrity flags but are not gate
members.

\paragraph{Experience store and isolation.}
The artifact README states that retrieval is disabled for the primary frozen
comparison and that experience data are omitted. When enabled, the verified
store uses append-only \texttt{data/experience.jsonl},
\texttt{data/experience.idx.npy}, and \texttt{data/experience.keys.txt}; the
retriever excludes the current task. The harness creates a timestamped run
folder, and the v5.1 driver uses one working directory per solver subprocess.
The historical pool records retrieval as inactive; the $n=21$ A/B is the only
reported retrieval comparison.

\paragraph{Models, evaluator, and budgets.}
The historical comparison uses \texttt{deepseek-v4-pro} and
\texttt{claude-opus-4-8}; the source annotation records up to 100 tool-calling
turns, 16,384 output tokens, a 7,200-s solver timeout, QE 7.5, and
\texttt{JOBSPEC-2026-07-05} for the frozen batch. The frozen artifact's harness
also defaults to a 7,200-s timeout and its v1 entry point defaults to 100 turns
and 16,384 output tokens; the v5.1 entry point defaults to 60 turns and 8,192
tokens unless overridden. This invocation-level difference remains an author
check for the historical v5.1 rows. The frozen score summary is
\texttt{results/frozen\_matched\_summary.json}; its recorded bootstrap seed is
20260817 with 100,000 samples.

\paragraph{Multiplicity.}
Figure~\ref{fig:compare} shows four paired M1+M2 cells for each model, an
eight-test family. Bonferroni at $\alpha=0.05$ gives cutoff $0.00625$.
The figure reports the deepseek bands+DOS result as $p=0.006$ rounded to three
decimals; the unrounded value and test script are not in the package, so its
corrected classification remains unresolved.

\bibliographystyle{plainnat}
\bibliography{refs/bibtex_core}

\end{document}